\pdfoutput=1
\documentclass[conference]{IEEEtran}
\IEEEoverridecommandlockouts
\usepackage[
  backend=biber,
  style=ieee,
  sorting=none,
  sortcites=true,
  giveninits=true,
  maxbibnames=3,
  minbibnames=1,
  doi=false,
  isbn=false,
  url=false
]{biblatex}
\DeclareFieldFormat[article,inproceedings,misc,online,report]{title}{\mkbibquote{#1\isdot}}
\usepackage{amsmath,amssymb,amsfonts}
\usepackage{algorithm,algorithmic}
\usepackage{graphicx}
\usepackage{multirow}
\usepackage{array}
\usepackage{placeins}
\usepackage{textcomp}
\usepackage{xcolor}
\usepackage{url}
\def\BibTeX{{\rm B\kern-.05em{\sc i\kern-.025em b}\kern-.08em
    T\kern-.1667em\lower.7ex\hbox{E}\kern-.125emX}}

\begin{document}
\title{SR-VLA: Stall-Reduced VLA Execution via Predicted-Input Lookahead and Selective KV Cache Repair%
\thanks{This work has been submitted to the IEEE for possible publication. Copyright may be transferred without notice, after which this version may no longer be accessible.}}
\author{
\IEEEauthorblockN{Tao Zhang\textsuperscript{1,2},\,%
Yi Zhong\textsuperscript{1,2*},\,%
Jinhao Ruan\textsuperscript{1,2},\,%
Yipeng Gao\textsuperscript{1,2},\,%
Linfeng Zhong\textsuperscript{1,2},\,%
Youming Yang\textsuperscript{1,2},\,%
and
Yuan Wang\textsuperscript{1,2,3*}}
\IEEEauthorblockA{\textsuperscript{1}School of Integrated Circuits, Peking University \\
\textsuperscript{2}Key Laboratory of Microelectronic Devices and Circuits, MPW Center, Peking University \\
\textsuperscript{3}Beijing Advanced Innovation Center for Integrated Circuits \\
*Corresponding Authors.}
}
\maketitle

\begin{abstract}
  Vision-language-action (VLA) policies close the perception--action loop of autonomous robots, but their inference blocks the robot at every action-chunk boundary. Asynchronous inference overlaps VLA inference with robot execution to reduce this stall. However, the inputs for the next decision do not yet exist when inference starts, so results computed in advance may mismatch the real inputs that arrive at the boundary, while correcting such mismatches by full recomputation puts the latency back on the critical path. Therefore, we propose SR-VLA, a stall-reduced execution architecture that addresses these issues. During execution, SR-VLA predicts the next-boundary observation and robot state from the committed actions, letting the frozen VLA precompute a speculative KV cache and action chunk. Once the real inputs arrive, SR-VLA checks them against the predictions and accepts the speculative action when they agree; otherwise, it repairs the cache by rebuilding only the high-risk visual tokens. On LIBERO, SR-VLA reduces mean stall from 128.11 to 47.95~ms, achieving $2.67\times$ and $1.27\times$ stall reductions over synchronous $\pi_{0.5}$ and OpenVLA-OFT with competitive success rates. Under target-motion perturbation, it keeps synchronous-level success with $1.52\times$--$1.93\times$ stall reductions.
  \end{abstract}

\section{Introduction}

\begin{figure}[!t]
  \centering
  \includegraphics[width=\columnwidth]{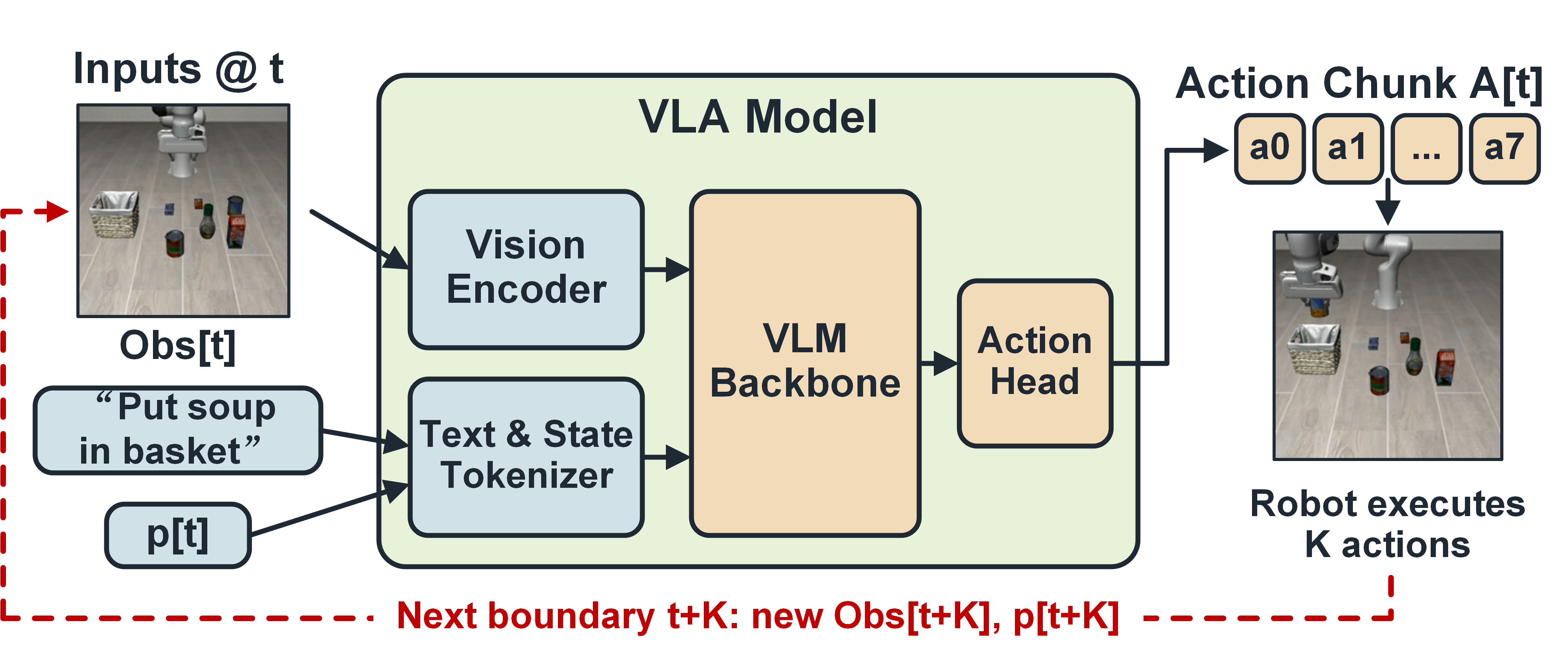}
  \caption{A VLA policy in the robot perception--action loop. At decision boundary $t$, the VLA model encodes the observation, language instruction, and robot state, and emits an action chunk of $K$ actions; the robot executes the chunk before the next decision at $t+K$.}
  \label{fig:vla_overview}
\end{figure}

Vision-language-action (VLA) models adapt large vision-language models into end-to-end robot policies that map visual observations, language instructions, and proprioceptive state to actions, generalizing across complex manipulation tasks \cite{rt2,openvla,pi0}. On an autonomous robot, the VLA policy sits inside the perception--action loop shown in Fig.~\ref{fig:vla_overview}: at each decision boundary, it consumes the latest sensor inputs and emits the next action chunk, which the controller then executes \cite{act,pi0,openvlaoft}. The responsiveness of the whole system is therefore bounded by how long the robot waits at each boundary.

\begin{figure*}[!t]
  \centering
  \makebox[\textwidth][c]{\includegraphics[width=1.04\textwidth,height=250pt]{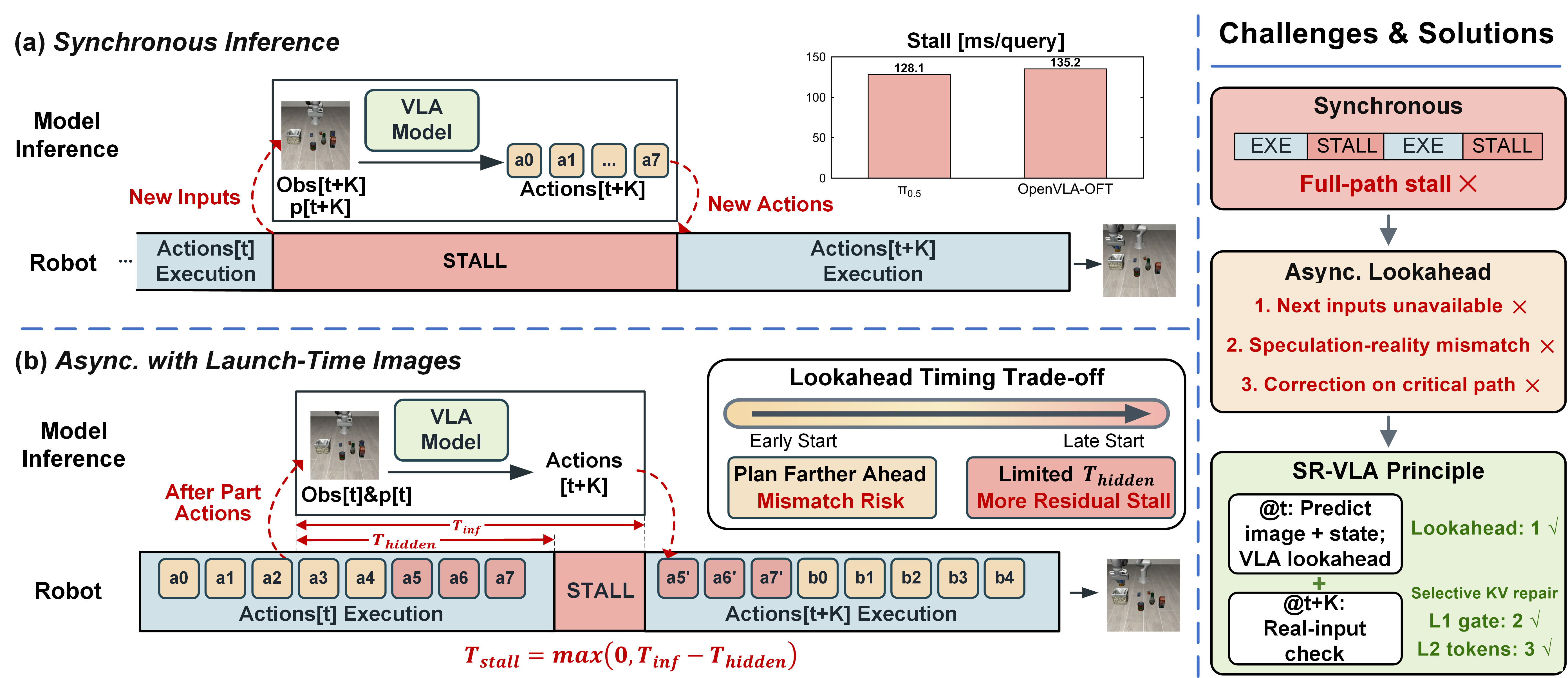}}
  \caption{Motivation of SR-VLA. (a) Synchronous inference exposes the full inference latency as robot stall. (b) Asynchronous inference with launch-time images trades residual stall against input staleness. Right: the challenges of asynchronous inference and the SR-VLA solutions, where lookahead addresses challenge 1 and the two levels (L1, L2) of selective KV repair address challenges 2 and 3.}
  \label{fig:execution_timing}
\end{figure*}

Under synchronous execution shown in Fig.~\ref{fig:execution_timing}(a), this wait is substantial: inference starts only after the next-boundary inputs arrive, so the robot stalls for the full inference latency. At a 30~Hz action rate, $\pi_{0.5}$~\cite{pi05} and OpenVLA-OFT~\cite{openvlaoft} stall for 128.1 and 135.2~ms per query, idling the robot for 43.5\% and 33.6\% of each control cycle. Model- and decoding-level accelerations shorten this stall, but each decision still waits for the inputs it conditions on. Asynchronous execution~\cite{smolvla,rtc} instead overlaps inference with the current chunk, but with launch-time inputs it faces the trade-off shown in Fig.~\ref{fig:execution_timing}(b): starting late leaves residual stall, whereas starting early plans from stale context. Future-input methods predict the execution-time context~\cite{vlash,futurertc,jetsonpi,pactvla}, yet they either consume it without checking it against the real boundary inputs or correct it only by choosing among stale candidates or rerunning the full model (Section~\ref{sec:related_work}).

Although prior asynchronous methods hide part of the inference latency by overlapping it with robot execution, fast and safe VLA inference still faces three challenges, as summarized in Fig.~\ref{fig:execution_timing}. \textbf{Firstly, the next-decision inputs are unavailable.} The observation and robot state that condition the next decision exist only after the current chunk ends. Direct inference with launch-time inputs would cut $\pi_{0.5}$ success from 96.4\% to 62.4\% on LIBERO \cite{libero}. \textbf{Secondly, speculation may mismatch reality.} Precomputed results may deviate from reality, e.g., when objects move. Even with accurate input prediction, unvalidated reuse reaches only 87.3\% success, and mismatched actions are unsafe for a physical robot. \textbf{Finally, correction lies on the critical path.} Correction starts only after the real inputs arrive, i.e., on the stall path. Fully recomputing at every detected mismatch still leaves stall.

To alleviate these problems, we propose SR-VLA, a stall-reduced VLA execution architecture with two components: predicted-input lookahead and selective KV cache repair. Firstly, the lookahead predicts the next-boundary image and robot state from the committed action chunk, letting the frozen VLA precompute a speculative KV cache and action chunk during execution. Secondly, the first level of selective KV repair checks the arriving real inputs against the predictions and task-critical conditions, and then accepts the speculative action or triggers repair. Finally, the second level of selective KV repair rebuilds only the high-risk visual tokens and reuses the remaining speculative KV entries. Our contributions are:

\begin{itemize}
    \item A speculate--validate--repair execution architecture that moves next-boundary input prediction and frozen-VLA lookahead off the critical path of the perception--action loop, precomputing a speculative KV cache and action chunk while the robot executes.

    \item A two-level selective KV cache repair mechanism that gates speculative reuse with real inputs and rebuilds only high-error and high-risk visual-token KV entries.

    \item An evaluation on LIBERO with frozen $\pi_{0.5}$ and OpenVLA-OFT, where SR-VLA achieves $2.67\times$ and $1.27\times$ stall reductions over the respective synchronous baselines at competitive success, and keeps synchronous-level success under target motion with $1.52\times$--$1.93\times$ stall reductions.
\end{itemize}

\section{Background and Related Work}

\subsection{VLA Execution Timing in the Perception--Action Loop} \label{sec:stall_metric}

In the loop of Fig.~\ref{fig:vla_overview}, a VLA policy maps the visual observation $o_t$, language instruction $g$, and robot state $p_t$ at decision boundary $t$ to an action chunk $A_t=[a_t,\ldots,a_{t+K-1}]$, whose $K$ steps the controller executes at rate $f$ before the next boundary $t+K$ \cite{openvla,openvlaoft,pi0}. Let $T_{\mathrm{exec}}=K/f$ denote their execution duration, $T_{\mathrm{inf}}$ the complete processing time needed to prepare the next action chunk, and $T_{\mathrm{hidden}}$ the portion actually covered by robot execution of $A_t$, as shown in Fig.~\ref{fig:execution_timing}(b). The exposed boundary stall is
\begin{equation}
T_{\mathrm{stall}}=\max(0,T_{\mathrm{inf}}-T_{\mathrm{hidden}}).
\label{eq:stall_paths}
\end{equation}
Because $o_{t+K}$ and $p_{t+K}$ exist only after $A_t$ completes, this data dependency places every decision boundary on the critical path of the loop. Synchronous inference therefore has $T_{\mathrm{hidden}}=0$, whereas asynchronous execution gains $T_{\mathrm{hidden}}>0$ only by starting from inputs available at launch time; its measured overlap need not span $T_{\mathrm{exec}}$.

\begin{figure*}[!t]
  \centering
  \includegraphics[width=\textwidth]{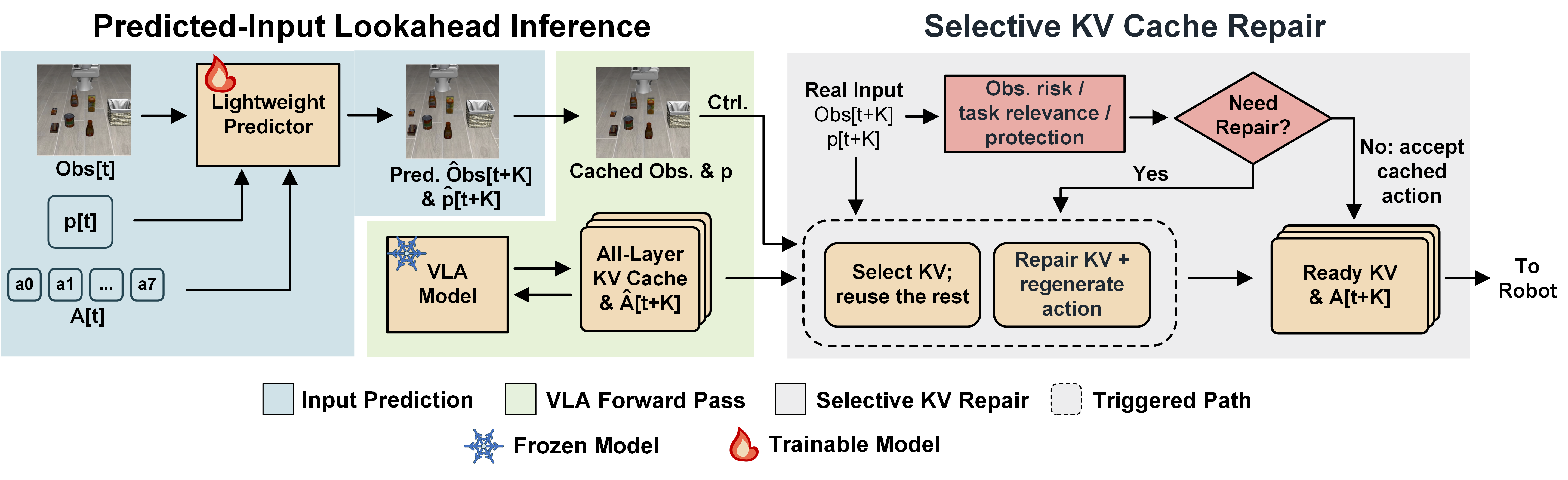}
  \caption{SR-VLA execution path. During committed $A_t$, predicted images and proprioceptive state drive the frozen VLA to construct a speculative KV cache and next action. At the target boundary, real inputs, task relevance, and protection conditions selectively repair the KV cache.}
  \label{fig:predict_and_repair_pipeline}
\end{figure*}

\subsection{Speculative Execution at Decision Boundaries} \label{sec:speculation}

During the execution of $A_t$, the controller has access to $o_t$, $p_t$, $g$, and the committed actions, while the next-boundary inputs remain unavailable. Given predicted inputs $\hat{o}_{t+K}$ and $\hat{p}_{t+K}$, lookahead execution can construct a speculative VLA state
\begin{equation}
\hat{\mathcal{S}}_{t+K}=F(\hat{o}_{t+K},g,\hat{p}_{t+K}),
\label{eq:spec_state}
\end{equation}
including the layer-wise states and KV entries required for the next decision. Processors face an analogous dependency when an operand is not yet ready: value prediction lets dependent instructions execute on a predicted operand, verifies the prediction before commit, and replays only the dependent work on a misprediction~\cite{valuepred}. Applied to the decision boundary, this discipline yields three requirements that mirror the challenges in Fig.~\ref{fig:execution_timing}: (1) input prediction and lookahead complete off the critical path; (2) the speculative result is validated against the real inputs before it is committed to the robot; and (3) correction is paid only at boundaries where the prediction fails and only for the invalidated state, keeping the expected correction cost on the critical path low.

\subsection{Related Work} \label{sec:related_work}

\textbf{Shortening $T_{\mathrm{inf}}$.} Model-level accelerations compress, quantize, prune, or skip layers of the VLA~\cite{tinyvla,qail,spvla,dyslvla,efficientvla,flashvla}; decoding-level methods generate actions through parallel or speculative decoding~\cite{parallelvla,specvla,kerv,realtimeflash}; and system-level methods pipeline inference or manage KV state on edge platforms~\cite{actionflow,keep}. VLA-Cache~\cite{vlacache} reuses tokens across adjacent real observations, and CacheBlend~\cite{cacheblend} recomputes only high-deviation tokens of reused KV caches in LLM serving. Like speculative decoding for language models~\cite{specdec,draftverify}, these methods speculate on or reuse outputs under an already available conditioning input, so each decision still waits for its inputs and $T_{\mathrm{hidden}}=0$.

\textbf{Increasing $T_{\mathrm{hidden}}$.} Asynchronous execution~\cite{smolvla,rtc} overlaps inference with the current chunk using launch-time inputs. Future-input methods address the resulting misalignment: VLASH~\cite{vlash} fine-tunes the base VLA with temporal offsets and rolls the robot state forward, FutureRTC~\cite{futurertc} predicts execution-time visual features and state for a frozen policy, and EagleVLA~\cite{jetsonpi} predicts compressed VLM representations and reruns the full VLM when its confidence drops, without predicting images or correcting the layer-wise KV cache. PACT-VLA~\cite{pactvla} selects among candidate chunks generated from the launch-time observation using a later pre-commit observation, and CheckVLA~\cite{checkvla} compares action-conditioned predictions with arriving observations to rewrite failing chunks. These methods address requirement (1) and, in some cases, (2). When a prediction is wrong, however, their correction is absent, limited to precomputed candidates, or a full re-inference, so requirement (3) and the resulting boundary cost are not part of their objective.

\textbf{Predictive world models.} Robot world models estimate future observations or latent states from current states and actions~\cite{gr1,robodreamer} for learning behavior or long-horizon planning rather than preparing the next VLA input at runtime. In contrast, SR-VLA addresses the three requirements together and treats the boundary cost of correction as a design objective.

\section{SR-VLA Architecture}
\subsection{Architecture Overview}
\setlength{\textfloatsep}{2pt plus 2pt minus 2pt}
\setlength{\dbltextfloatsep}{2pt plus 2pt minus 2pt}
\setlength{\floatsep}{2pt plus 2pt minus 2pt}

As illustrated in Fig.~\ref{fig:predict_and_repair_pipeline}, SR-VLA follows the three requirements of Section~\ref{sec:speculation} by separating overlapable lookahead from post-arrival validation. During the execution of $A_t$, a trainable predictor uses $o_t,p_t,A_t$ to predict the next-boundary image $\hat{o}_{t+K}$ and state $\hat p_{t+K}$. The image passes through the original visual encoder, and the frozen VLA uses both predictions to produce $\hat{A}_{t+K}$ and its all-layer KV cache $\hat{\mathcal{K}}_{t+K}$. This speculative pair $(\hat{\mathcal{K}}_{t+K},\hat{A}_{t+K})$ is stored and will not influence the current executing action chunk.

When the current action chunk is consumed by the robot, the real observation $o_{t+K}$ and proprioceptive state $p_{t+K}$ arrive, and SR-VLA decides whether repair is required. The trigger combines an observation-driven gate, which checks the patch mismatch between predicted and real inputs, and a task-protection gate, which enforces mandatory updates. If neither gate is active, $(\hat{\mathcal{K}}_{t+K},\hat{A}_{t+K})$ is accepted and the predicted action chunk is forwarded to the robot. Otherwise, SR-VLA recomputes the affected visual-token KV entries from the real inputs and replaces the speculative pair with a repaired KV cache $\mathcal{K}^{\mathrm{rep}}_{t+K}$ and a regenerated action $A^{\mathrm{rep}}_{t+K}$ before any action is sent to the robot.

Input prediction and frozen-VLA lookahead form the prediction path, and post-arrival processing forms the real-input path, so the complete processing time in Eq.~(\ref{eq:stall_paths}) becomes
\begin{equation}
T_{\mathrm{inf}}=T_{\mathrm{pred}}+T_{\mathrm{real}}.
\label{eq:latency_sr_vla}
\end{equation}
$T_{\mathrm{pred}}$ ends when the speculative KV cache and action chunk are ready. Only this path can overlap execution of $A_t$, giving $0\leq T_{\mathrm{hidden}}\leq\min(T_{\mathrm{pred}},T_{\mathrm{exec}})$. $T_{\mathrm{real}}$ begins once the real boundary inputs and the speculative pair are available, and covers image encoding, trigger evaluation, and either acceptance or selective KV repair and action regeneration; it therefore lies entirely on the critical path. Splitting this path by whether repair is triggered and averaging over boundaries gives
\begin{equation}
\mathbb{E}[T_{\mathrm{real}}]=\bar{T}_{\mathrm{val}}+r\,\bar{T}_{\mathrm{rep}},
\label{eq:real_path}
\end{equation}
where $T_{\mathrm{val}}$ covers image encoding and trigger evaluation at every boundary, $T_{\mathrm{rep}}$ covers selective KV repair and action regeneration at triggered boundaries, and $r$ is the repair rate. Requirement (3) thus has two factors, which the two levels of selective repair target separately. Besides enforcing requirement (2), the first level lowers $r$ by accepting the speculative pair whenever the real inputs confirm it, and the second level lowers $\bar{T}_{\mathrm{rep}}$ by rebuilding only the visual-token KV entries invalidated by the real inputs and reusing the rest (Section~\ref{sec:selective_repair}).

\begin{figure}[!t]
  \centering
  \makebox[\columnwidth][c]{\includegraphics[width=1.04\columnwidth,height=112pt]{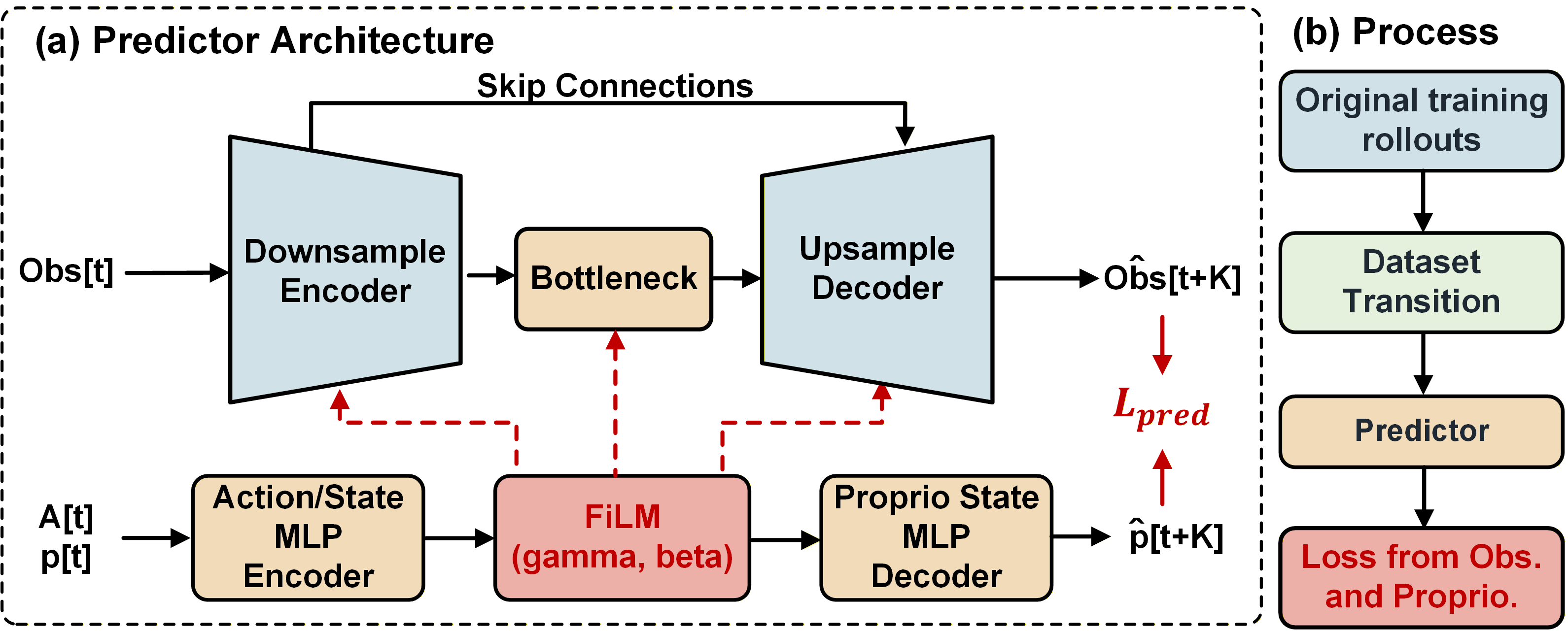}}
  \caption{Joint next-boundary image and proprioceptive-state prediction, conditioned on the current image, robot state, and committed action chunk.}
  \label{fig:next_boundary_predictor}
\end{figure}

\subsection{Action-Conditioned Next-Boundary Input Predictor} \label{sec:predictor}

After $A_t$ is produced, the predictor receives the current visual observation $o_t$, robot state $p_t$, and the executing action chunk $A_t$. As shown in Fig.~\ref{fig:next_boundary_predictor}, it jointly predicts the visual observation and proprioceptive state at the next decision boundary:
\begin{equation}
(\hat{o}_{t+K},\hat{p}_{t+K})=P(o_t,p_t,A_t;\theta_P).
\label{eq:predictor}
\end{equation}
Here, $P$ denotes the action-conditioned next-boundary predictor, and $\theta_P$ denotes its trainable parameters. The predictor advances one execution interval, from $t$ to the boundary after $K$ executed control steps.

The predictor comprises a proprioceptive-state branch and an image branch, which produce the aligned $\hat{p}_{t+K}$ and $\hat{o}_{t+K}$, respectively, for the same next-boundary VLA input. An action/state MLP jointly encodes $A_t$ and $p_t$ into a shared representation. The proprioceptive-state branch uses an MLP decoder to map this representation to $\hat{p}_{t+K}$. The image branch takes $o_t$ as input and predicts $\hat{o}_{t+K}$ using a compact U-Net with a downsampling encoder, a bottleneck, and a symmetric upsampling decoder. It targets only the next decision boundary rather than rolling out a long-horizon world model. The shared action/state representation also conditions the image branch through feature-wise linear modulation (FiLM). At stage $l$, the image features are modulated as
\begin{equation}
  \tilde{h}_l=\gamma_l(A_t,p_t)\odot h_l + \beta_l(A_t,p_t),
\label{eq:film}
\end{equation}
where $h_l$ and $\tilde{h}_l$ denote the image features before and after modulation at stage $l$, respectively. The scaling parameters $\gamma_l$ and shifting parameters $\beta_l$ are generated by the action/state MLP, and $\odot$ denotes element-wise multiplication. This modulation conditions the predicted spatial change on the executing action chunk and current robot state.

\begin{table}[!t]
  \centering
  \setlength{\abovecaptionskip}{3pt}
  \caption{Next-boundary predictor ablation on LIBERO-Spatial with frozen $\pi_{0.5}$; boundary repair is disabled.}
  \label{tab:predictor_ablation}
  \scriptsize
  \setlength{\tabcolsep}{1.4pt}
  \renewcommand{\arraystretch}{1.08}
  \resizebox{\columnwidth}{!}{%
  \begin{tabular}{lccccc}
  \hline
  \multirow{2}{*}{Setting} &
  \multicolumn{3}{c}{Patch cosine similarity $\uparrow$} &
  \multirow{2}{*}{\shortstack[c]{Proprio.\\MSE $\downarrow$}} &
  \multirow{2}{*}{\shortstack[c]{Closed-loop\\success (\%) $\uparrow$}} \\
  \cline{2-4}
  & $\tau_d{=}0.5$ & $\tau_d{=}0.95$ & $\tau_d{=}1.0$ & & \\
  \hline
  \textbf{Sync} ($\pi_{0.5}$, real input) & 1.000 & 1.000 & 1.000 & 0.000 & 97.8 \\
  \textbf{Async} (w/o prediction) & -0.037 & 0.394 & 0.856 & 0.069 & 45.2 \\
  \hline
  \textbf{Predictor} & 0.793 & 0.854 & 0.963 & 0.004 & 87.6 \\
  \hline
  \multicolumn{6}{l}{\textbf{Ablation variants}} \\
  w/o FiLM & 0.551 & 0.697 & 0.924 & 0.004 & 70.6 \\
  w/o Proprio. & 0.790 & 0.852 & 0.963 & 0.069 & 68.0 \\
  w/o Action Conditioning & 0.566 & 0.709 & 0.928 & 0.006 & 62.2 \\
  \hline
  \end{tabular}}
  \vspace{1pt}
  \parbox{0.98\columnwidth}{\scriptsize
  \emph{Note:} Cosine similarity is averaged over patches whose current-to-next real-patch cosine similarity is $\leq \tau_d$. $\tau_d=1.0$ includes all patches. The Sync row uses real inputs.}
\end{table}

Training pairs are generated offline from ground-truth rollouts under the original training setup~\cite{pi05,openvlaoft}. The dataset needs to contain aligned $K$-step transitions from boundary $t$ to boundary $t+K$. Therefore, the current observation, current proprioceptive state, and action chunk form the predictor-side context, while $o_{t+K}$ and $p_{t+K}$ supervise the two next-boundary outputs. We train the predictor with two complementary error terms:
\begin{equation}
\mathcal{L}_{\mathrm{pred}}
=\mathcal{L}_{\mathrm{img}}
+\lambda_p\mathcal{L}_{\mathrm{prop}},
\label{eq:predictor_loss}
\end{equation}
where $\mathcal{L}_{\mathrm{img}}=\mathbb{E}_{u}[\sqrt{(\hat{o}_{t+K,u}-o_{t+K,u})^2+\epsilon^2}]$ is a Charbonnier loss over all image elements $u$, with $\epsilon=10^{-6}$, and $\mathcal{L}_{\mathrm{prop}}=\mathbb{E}[\|\hat{p}_{t+K}-p_{t+K}\|_2^2]$ is the proprioceptive-state MSE. The coefficient is set to $\lambda_p=0.1$. The predictor itself contains 10.08M parameters and requires 51.52 GFLOPs per inference.

We train the predictor on 13622 samples for 100k steps using AdamW (batch size 8, learning rate $10^{-4}$, and weight decay $10^{-4}$), and use a disjoint set of 960 samples for validation. To assess the contributions of action conditioning, FiLM, and the proprioceptive-state branch, we conduct the component ablations in Table~\ref{tab:predictor_ablation} on LIBERO-Spatial with KV repair disabled. We choose this suite because launch-time inputs degrade success most on it among the four suites, and Section~\ref{sec:stall_breakdown} repeats the analysis over all four suites. All variants use the same frozen VLA backbone~\cite{pi05}, and Async (w/o prediction) directly uses launch-time $o_t,p_t$ for the next-boundary action. Experimental results reveal that prediction could narrow the success gap to the real-input synchronous baseline, raising closed-loop success from 45.2\% without prediction to 87.6\%. Removing FiLM or action conditioning degrades the image similarity compared to the full predictor, while removing the state branch leaves image similarity nearly unchanged but worsens proprioceptive accuracy and lowers success rate to 68.0\%. 

The remaining gap between the predicted results and real inputs motivates a boundary validation for this predictor. Section~\ref{sec:selective_repair} therefore examines selective repair.

\subsection{Boundary Validation and Selective KV Cache Repair} \label{sec:selective_repair}

The arriving real inputs may invalidate only part of the speculative KV--action pair. Algorithm~\ref{alg:selective_repair} implements two levels of selectivity: decide whether repair is needed, then select which visual-token KV entries to rebuild.

\begin{algorithm}[!t]
\caption{Selective KV Cache Repair at a Decision Boundary}
\label{alg:selective_repair}
\begin{algorithmic}[1]
\setlength{\itemsep}{0.7pt}
\footnotesize
\REQUIRE lookahead $\hat{A},\hat{p},\hat{z},\hat{W},\hat{\mathcal{K}}$
\REQUIRE real observation/state $o,p$, instruction $g$, counter $c$
\STATE $z\leftarrow\operatorname{VisualTokens}(o)$
\STATE $\boldsymbol{\rho}\leftarrow
\operatorname{Norm}(\operatorname*{mean}_{q\in Q_g\cup Q_a}
\hat{W}_{q,\mathcal V})$
\STATE $e_i\leftarrow 1-\operatorname{CosSim}(\hat z_i,z_i),
\quad \forall i\in\mathcal V$
\STATE $r_i\leftarrow \rho_i e_i,
\quad \forall i\in\mathcal V$
\STATE $R_{\mathrm{img}}\leftarrow\max_{i\in\mathcal V}r_i$
\STATE $q_{\mathrm{obs}}\leftarrow(R_{\mathrm{img}}>\tau_{\mathrm{img}})\lor(\operatorname{RMSE}(\hat{p},p)>\tau_p)$
\STATE $q_{\mathrm{task}}\leftarrow\operatorname{GripCritical}(\hat{A},p)\lor(c\ge C_{\max})$
\IF{$\neg(q_{\mathrm{obs}}\lor q_{\mathrm{task}})$}
  \RETURN $\hat{A},\hat{\mathcal{K}},c+1$
\ENDIF
\STATE $\mathcal S_{\mathrm{upd}}\leftarrow
\operatorname{TopK}(\mathbf e,k_e)\cup
\operatorname{TopRatio}(\mathbf r,\alpha_r)$
\STATE $\mathcal{S}_{\mathrm{reuse}}\leftarrow\mathcal V\setminus\mathcal{S}_{\mathrm{upd}}$
\STATE $\mathcal{K}\leftarrow\operatorname{RepairKV}(z,p,g,\hat{\mathcal{K}},\mathcal{S}_{\mathrm{upd}},\mathcal{S}_{\mathrm{reuse}})$
\STATE $A\leftarrow\operatorname{RegenerateAction}(\mathcal{K},p,g)$
\RETURN $A,\mathcal{K},0$
\end{algorithmic}
\end{algorithm}

\begin{figure}[!t]
  \centering
  \includegraphics[width=\columnwidth,keepaspectratio]{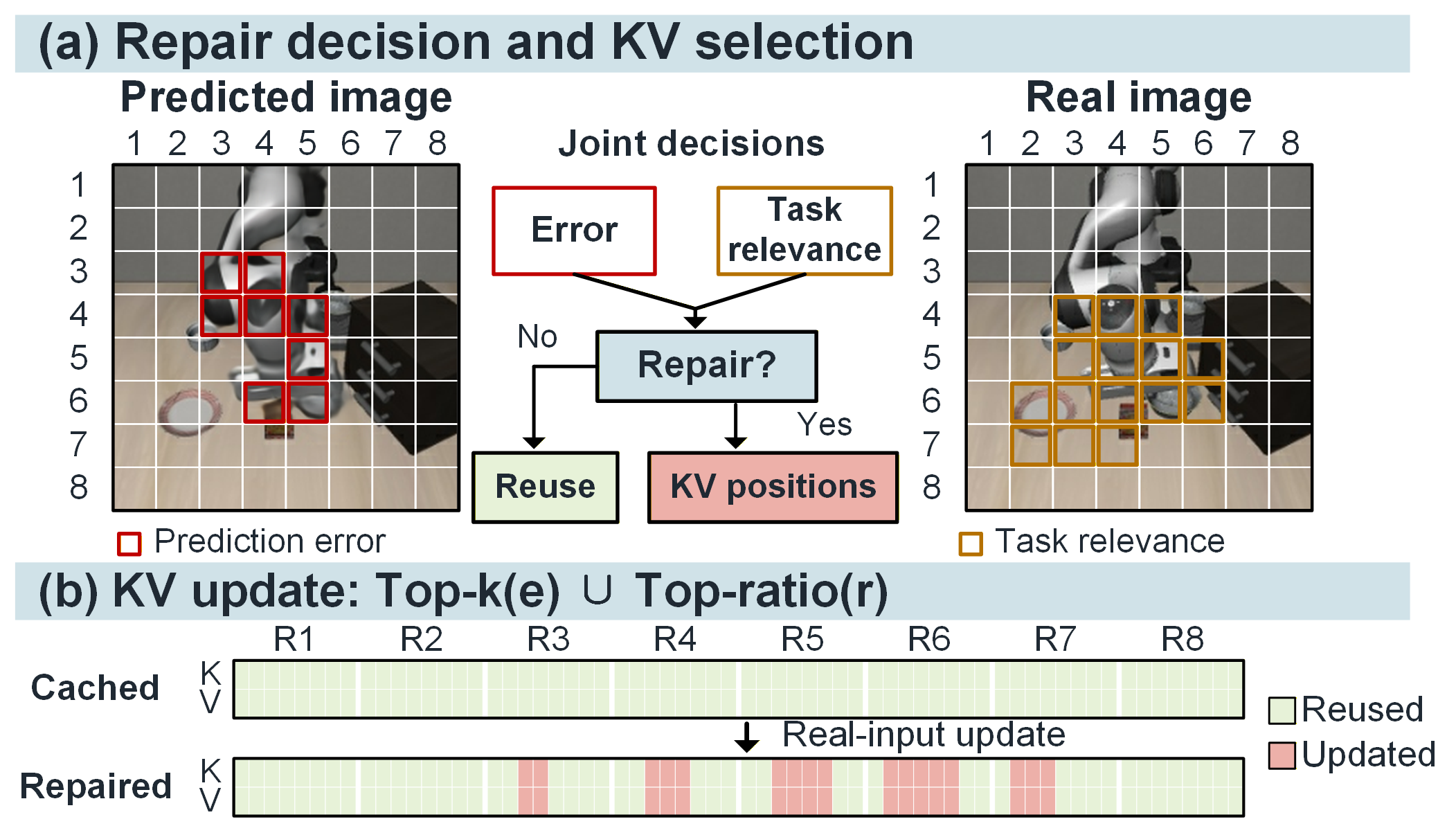}
  \caption{Selective KV repair schematic. (a) predict-real error and task relevance on image patches. (b) Visual KV replacement.}
  \label{fig:selective_kv_repair}
\end{figure}

Whether to repair is decided by either an input-driven gate or a task-protection gate. On the image side, visual tokens differ in temporal variation and task relevance~\cite{vlacache,cogvla,specprunevla}, motivating the patchwise validation illustrated in Fig.~\ref{fig:selective_kv_repair}(a). The arriving image is encoded into tokens $z$ and compared with the saved prediction $\hat z$, yielding the cosine error $e_i=1-\operatorname{CosSim}(\hat z_i,z_i)$. Task relevance $\rho_i$ is recovered from the normalized saved lookahead attention $\hat W$ assigned by language queries $Q_g$ and action queries $Q_a$ to patch $i$. Reusing $\hat W$ provides this task context without a complete real-image VLA pass before gating. The patchwise risk $r_i=\rho_i e_i$ combines the task relevance $\rho_i$ with the prediction cosine error $e_i$. The highest patch risk, $R_{\mathrm{img}}=\max_i r_i$, serves as the image-side gate score. Alongside the image-side assessment, SR-VLA measures proprioceptive mismatch using $\operatorname{RMSE}(\hat p,p)$. The input-driven gate activates when $R_{\mathrm{img}}>\tau_{\mathrm{img}}$ or $\operatorname{RMSE}(\hat p,p)>\tau_p$, allowing either input modality to trigger repair. Here, $\tau_{\mathrm{img}}$ and $\tau_p$ are thresholds. Independently, the task-protection gate activates when the speculative action chunk $\hat A$ predicts a gripper opening or closing operation, or after $C_{\max}$ consecutive speculative reuses. The counter thus bounds reliance on prediction: at most $C_{\max}$ consecutive action chunks are committed from predicted inputs before a repair writes the real inputs into the KV cache. If either gate is active, SR-VLA proceeds to selective KV repair; otherwise, it accepts predicted $(\hat{\mathcal K},\hat A)$ and increments the reuse counter.

Triggered repair applies the second level of selectivity to KV reuse. SR-VLA captures the largest raw changes with $\operatorname{TopK}(\mathbf{e},k_e)$, and prioritizes relevance-weighted risk with $\operatorname{TopRatio}(\mathbf{r},\alpha_r)$. Here, $k_e$ specifies the number of highest-error visual tokens selected per view, and $\alpha_r$ specifies the fraction selected by the risk $\mathbf{r}$. Their union forms the mandatory KV update set $\mathcal S_{\mathrm{upd}}$, as illustrated in Fig.~\ref{fig:selective_kv_repair}(b). Besides, the predicted proprioceptive input $\hat{p}$ is replaced by the real one $p$. $\operatorname{RepairKV}$ then runs a layer-by-layer partial forward pass from the real inputs: at layer $l$, only tokens in $\mathcal S_{\mathrm{upd}}$ compute queries, keys, and values, their keys and values overwrite the speculative entries in place, and only their queries attend over all keys:
\begin{equation}
h^{l+1}_{\mathcal S_{\mathrm{upd}}}=\operatorname{Layer}_l\big(h^{l}_{\mathcal S_{\mathrm{upd}}};\,[K^l_{\mathcal S_{\mathrm{upd}}};\hat K^l_{\mathcal S_{\mathrm{reuse}}}],[V^l_{\mathcal S_{\mathrm{upd}}};\hat V^l_{\mathcal S_{\mathrm{reuse}}}]\big).
\label{eq:repair_kv}
\end{equation}
This is the sparse-query computation of KV-cached decoding applied to scattered positions, as in selective KV recomputation for LLM serving~\cite{cacheblend}. Positions are unchanged, so no re-encoding is needed, and every per-layer cost scales with $|\mathcal S_{\mathrm{upd}}|$, yielding the cost model after Eq.~(\ref{eq:real_path}). Reused entries remain approximate, since the deeper-layer entries of an unchanged token attended to predicted versions of changed tokens; Top-$k$ error removes the largest such deviations, and Top-ratio risk refreshes the tokens most attended by action and language queries. Finally, the frozen VLA regenerates $A$ using the repaired KV, replacing $\hat{A}$. 

\begin{table}[!t]
\centering
\setlength{\abovecaptionskip}{3pt}
  \caption{Two-level repair-policy ablation with the predictor and frozen $\pi_{0.5}$ backbone fixed at $K=5$.}
\label{tab:repair_policy_ablation}
\scriptsize
\setlength{\tabcolsep}{1.8pt}
\renewcommand{\arraystretch}{1.08}
\resizebox{\columnwidth}{!}{%
\begin{tabular}{lcccc}
\hline
Repair setting & \shortstack{Task\\success (\%) $\uparrow$} & \shortstack{Action\\dev. $\downarrow$} & \shortstack{Repair\\rate (\%)} & \shortstack{Updated KV\\(all-query avg., \%)} \\
\hline
Full policy & 97.8 & 0.166 & 31.68 & 27.37 \\
  w/o input-driven gate & 91.4 & 0.211 & 20.2 & 17.1 \\
w/o task-protection gate & 94.6 & 0.176 & 30.8 & 25.4 \\
w/o Top-$k$ error & 91.0 & 0.246 & 36.3 & 0.5 \\
w/o Top-ratio risk & 95.4 & 0.163 & 34.1 & 27.6 \\
\hline
\end{tabular}}
\vspace{1pt}
\parbox{\columnwidth}{\scriptsize \emph{Note:} Spatial suite-wise settings, parameters used in evaluation: $\tau_{\mathrm{img}}=0.58$, $\tau_p=0.40$, $C_{\max}=27$, $k_e=220$ tokens/view, $\alpha_r=0.015$.}
\end{table}

Table~\ref{tab:repair_policy_ablation} evaluates the two selectivity levels on LIBERO-Spatial with the predictor, frozen $\pi_{0.5}$ backbone, execution chunk length $K=5$, and all other settings fixed. The full policy matches the synchronous baseline at 97.8\% success with a 31.68\% repair rate. Removing either repair gate lowers success, as does excluding Top-$k$ error or Top-ratio risk from token selection, so control quality depends on both when repair runs and which tokens it updates, rather than on repair frequency or update volume alone.

\section{Evaluation} \label{sec:evaluation}

\begin{figure*}[!t]
\centering
\begin{minipage}[b]{0.325\textwidth}
\centering
\includegraphics[width=\linewidth,height=124pt]{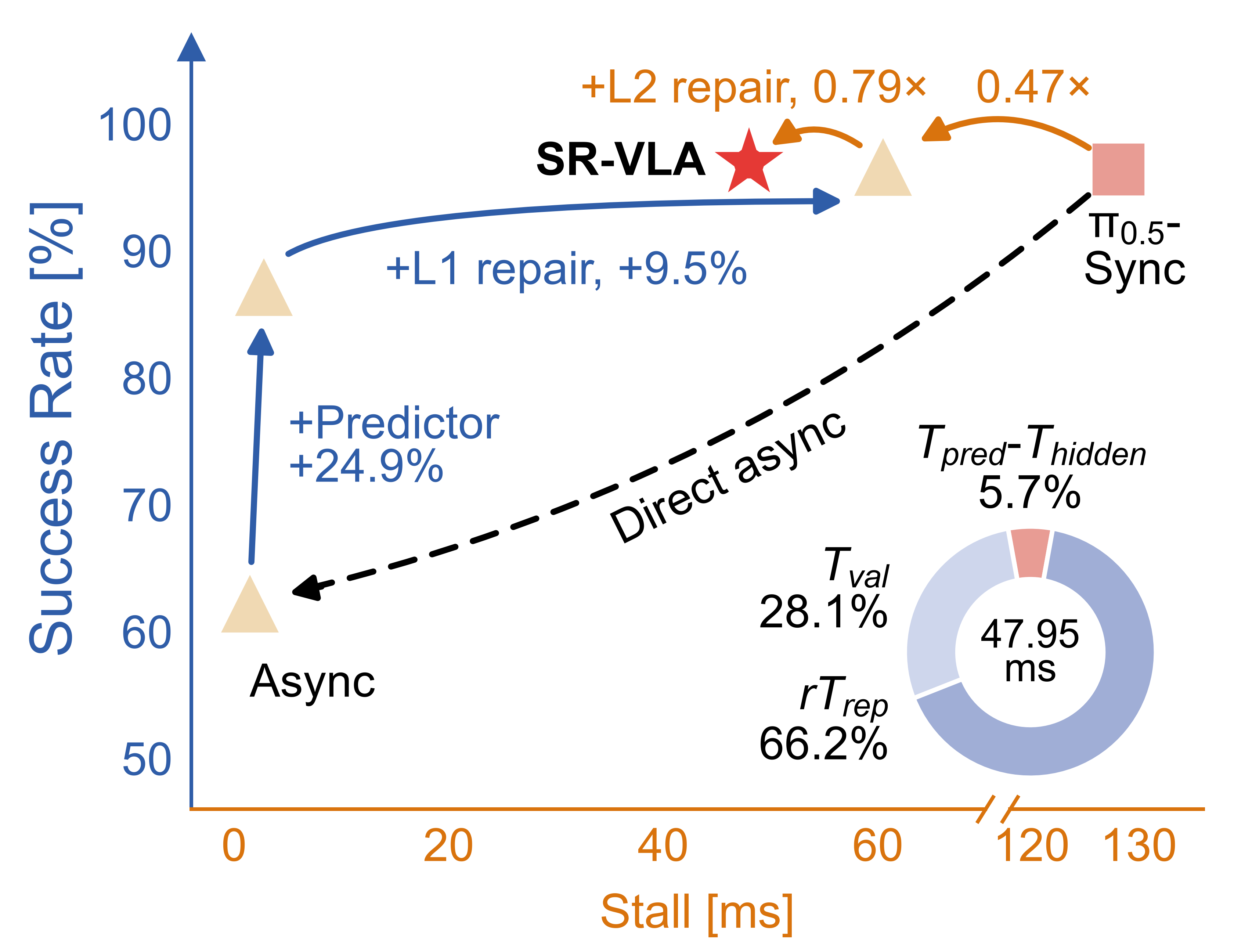}
\par\vspace{-1.0ex}{\footnotesize\textbf{(a)} Stall breakdown by component}
\end{minipage}\hfill
\begin{minipage}[b]{0.325\textwidth}
\centering
\includegraphics[width=\linewidth,height=120pt]{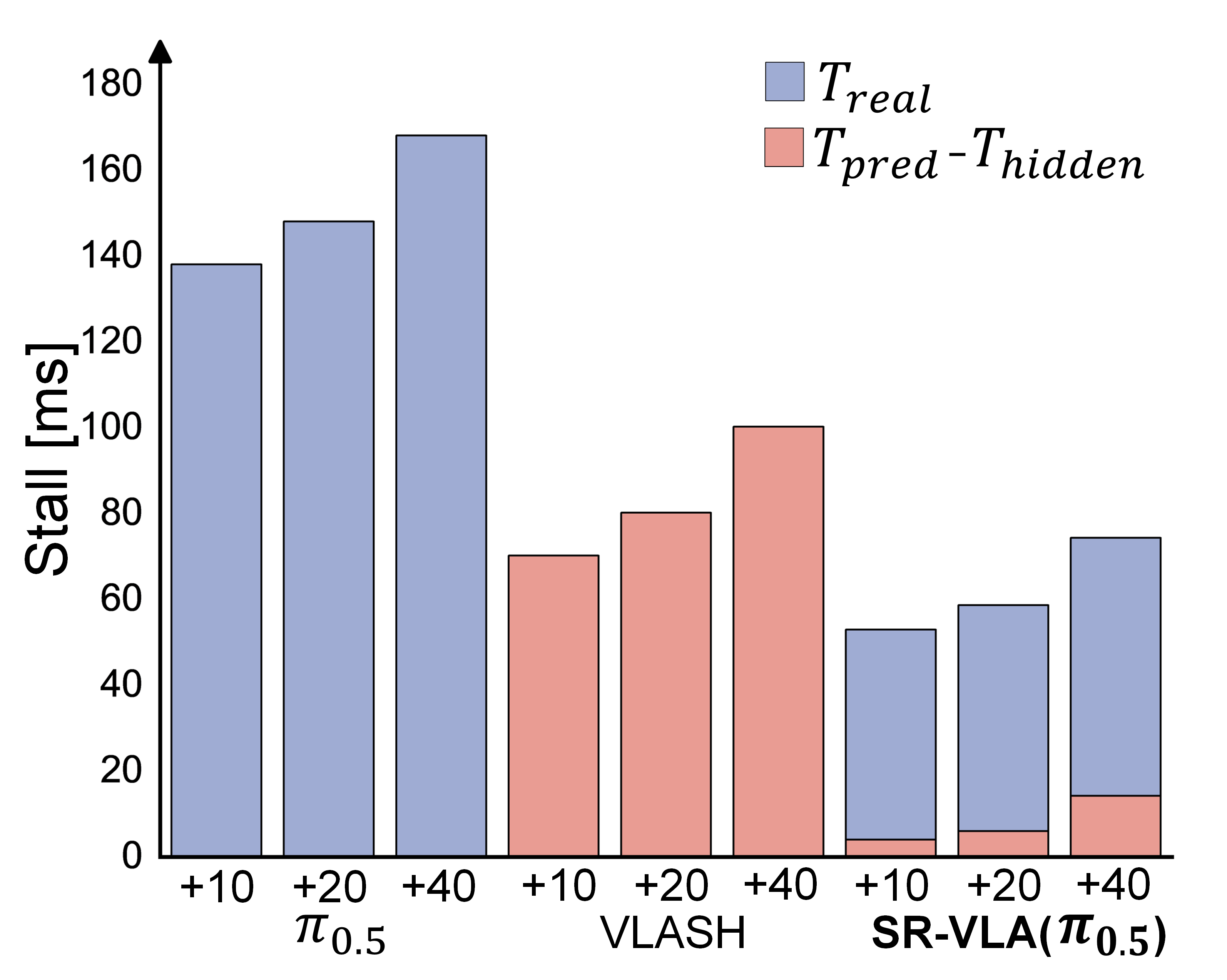}
\par\vspace{-1.0ex}{\footnotesize\textbf{(b)} Injected-delay sensitivity}
\end{minipage}\hfill
\begin{minipage}[b]{0.325\textwidth}
\centering
\includegraphics[width=\linewidth,height=116pt]{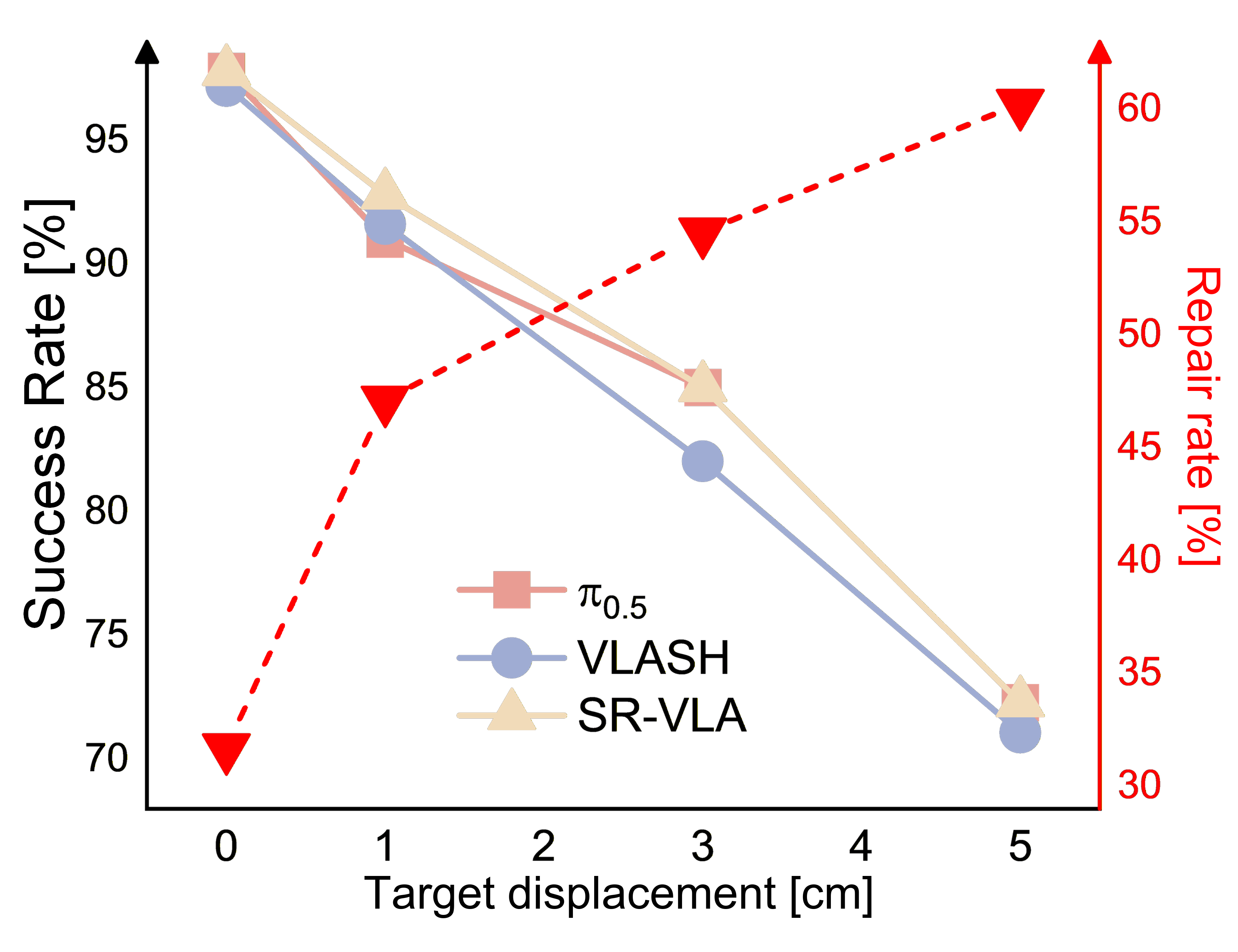}
\par\vspace{-1.0ex}{\footnotesize\textbf{(c)} Target-motion perturbation}
\end{minipage}
\caption{Stall breakdown, delay sensitivity, and target-motion perturbation results. (a) Stall and success of $\pi_{0.5}$ averaged over the four LIBERO suites as SR-VLA components are added. The inset breaks down the SR-VLA stall into exposed prediction, validation, and repair. (b) Mean stall under injected delay, split into exposed prediction and real-input computation. (c) LIBERO-Spatial task success and SR-VLA repair rate under target displacement.}
\label{fig:latency_and_target_motion_results}
\end{figure*}

We organize the evaluation around four questions. \textbf{Q1:} Does SR-VLA reduce decision-boundary stall while preserving task success across VLA backbones (Section~\ref{sec:eval_e2e})? \textbf{Q2:} How do lookahead, validation, and selective repair each address the three challenges in Fig.~\ref{fig:execution_timing} (Section~\ref{sec:stall_breakdown})? \textbf{Q3:} Does the stall advantage persist when inference slows down on deployment platforms (Section~\ref{sec:delay_sensitivity})? \textbf{Q4:} Does boundary validation respond to real input mismatches when the environment changes during execution (Section~\ref{sec:env_dynamics})?

\subsection{Experimental Setup}

We evaluate SR-VLA on the four LIBERO suites (Spatial, Object, Goal, and Long) with two state-of-the-art VLA backbones, $\pi_{0.5}$~\cite{pi05} and OpenVLA-OFT~\cite{openvlaoft}, whose checkpoints stay frozen and are shared with their synchronous baselines. SR-VLA uses only each backbone's input formats, visual-token embeddings, attention weights, and KV cache. We keep each model's native chunk length: $\pi_{0.5}$ and methods built on it use $K=5$~\cite{pi05,vlash}, whereas OpenVLA-OFT and methods built on it use $K=8$~\cite{vlacache,openvlaoft}. All measurements are collected on a 40GB A100 GPU at a 30~Hz action rate, so each chunk executes for $T_{\mathrm{exec}}=K/30$~s.

\begin{table}[!t]
  \centering
  \caption{Task comparisons on LIBERO with the $\pi_{0.5}$ backbone ($K=5$).}
  \label{tab:overall_pi05}
  \vspace{0.3ex}
  {\scriptsize
  \setlength{\tabcolsep}{0.80pt}
  \renewcommand{\arraystretch}{1.40}
  \begin{tabular}{@{}
  >{\centering\arraybackslash}m{0.235\columnwidth}
  >{\centering\arraybackslash}m{0.24\columnwidth}
  >{\centering\arraybackslash}m{0.065\columnwidth}
  >{\centering\arraybackslash}m{0.08\columnwidth}
  >{\centering\arraybackslash}m{0.085\columnwidth}
  >{\centering\arraybackslash}m{0.10\columnwidth}
  >{\centering\arraybackslash}m{0.10\columnwidth}
  @{}}
  \hline
  Method & \multicolumn{2}{c}{%
    \begin{tabular}{@{}>{\centering\arraybackslash}m{0.24\columnwidth}>{\centering\arraybackslash}m{0.065\columnwidth}@{}}
    \multicolumn{2}{c}{Success (\%) $\uparrow$} \\
    \cline{1-2}
    S/O/G/L & Avg. \\
    \end{tabular}%
  } & \shortstack{Stall/\\Action\\(ms) $\downarrow$} & \shortstack{Stall\\Mean\\(ms) $\downarrow$} & \scalebox{0.78}{\shortstack{Residual\\Stall\\($\times$) $\downarrow$}} & \scalebox{0.78}{\shortstack{Stall\\Reduction\\($\times$) $\uparrow$}} \\
  \hline
  \scalebox{0.88}{$\pi_{0.5}$-Sync~\cite{pi05}} & \scalebox{0.88}{\textbf{97.8}/97.4/97.0/93.4} & 96.4 & 25.62 & 128.11 & 1.00 & 1.00 \\
  \scalebox{0.88}{VLASH~\cite{vlash} ($d{=}1$)} & \scalebox{0.88}{\textbf{97.8}/99.4/97.0/93.4} & \textbf{96.9} & 18.72 & 93.58 & 0.73 & 1.37 \\
  \scalebox{0.88}{VLASH~\cite{vlash} ($d{=}2$)} & \scalebox{0.88}{97.2/\textbf{99.6}/95.6/\textbf{94.0}} & 96.6 & 12.05 & 60.25 & 0.47 & 2.13 \\
  \color{gray}\scalebox{0.88}{VLASH~\cite{vlash} ($d{=}3$)} & \color{gray}\scalebox{0.88}{94.4/98.8/93.3/91.9} & \color{gray}94.6 & \color{gray}5.38 & \color{gray}26.92 & \color{gray}0.21 & \color{gray}4.76 \\
  \hline
  \scalebox{0.88}{SR-VLA} & \scalebox{0.88}{\textbf{97.8}/99.2/\textbf{97.2}/93.4} & \textbf{96.9} & \textbf{9.59} & \textbf{47.95} & \textbf{0.37} & \textbf{2.67} \\
  \hline
  \end{tabular}
  \par\vspace{2pt}
  \parbox{0.96\columnwidth}{\raggedright\scriptsize S/O/G/L: Spatial/Object/Goal/Long; $d$: VLASH offset in actions ($d{=}2$ in Fig.~\ref{fig:latency_and_target_motion_results}). Residual Stall: mean stall per query over that of $\pi_{0.5}$-Sync; Stall Reduction: its inverse. \textbf{Bold}: best among settings with Avg. success $\geq$ $\pi_{0.5}$-Sync (gray: below).\par}
  }
\end{table}

\begin{table}[!t]
  \centering
  \caption{Task comparisons on LIBERO with the OpenVLA-OFT backbone ($K=8$).}
  \label{tab:overall_oft}
  \vspace{0.3ex}
  {\scriptsize
  \setlength{\tabcolsep}{0.80pt}
  \renewcommand{\arraystretch}{1.40}
  \begin{tabular}{@{}
  >{\centering\arraybackslash}m{0.235\columnwidth}
  >{\centering\arraybackslash}m{0.24\columnwidth}
  >{\centering\arraybackslash}m{0.065\columnwidth}
  >{\centering\arraybackslash}m{0.08\columnwidth}
  >{\centering\arraybackslash}m{0.085\columnwidth}
  >{\centering\arraybackslash}m{0.10\columnwidth}
  >{\centering\arraybackslash}m{0.10\columnwidth}
  @{}}
  \hline
  Method & \multicolumn{2}{c}{%
    \begin{tabular}{@{}>{\centering\arraybackslash}m{0.24\columnwidth}>{\centering\arraybackslash}m{0.065\columnwidth}@{}}
    \multicolumn{2}{c}{Success (\%) $\uparrow$} \\
    \cline{1-2}
    S/O/G/L & Avg. \\
    \end{tabular}%
  } & \shortstack{Stall/\\Action\\(ms) $\downarrow$} & \shortstack{Stall\\Mean\\(ms) $\downarrow$} & \scalebox{0.78}{\shortstack{Residual\\Stall\\($\times$) $\downarrow$}} & \scalebox{0.78}{\shortstack{Stall\\Reduction\\($\times$) $\uparrow$}} \\
  \hline
  \scalebox{0.88}{OFT-Sync~\cite{openvlaoft}} & \scalebox{0.88}{97.6/\textbf{98.4}/97.9/94.5} & 97.1 & 16.90 & 135.18 & 1.00 & 1.00 \\
  \scalebox{0.88}{VLA-Cache~\cite{vlacache}} & \scalebox{0.88}{98.3/97.5/\textbf{98.3}/\textbf{95.4}} & \textbf{97.4} & 15.24 & 121.88 & 0.90 & 1.11 \\
  \hline
  \scalebox{0.88}{SR-VLA} & \scalebox{0.88}{\textbf{98.8}/98.2/97.6/94.2} & 97.2 & \textbf{13.27} & \textbf{106.19} & \textbf{0.79} & \textbf{1.27} \\
  \hline
  \end{tabular}
  \par\vspace{2pt}
  \parbox{0.96\columnwidth}{\raggedright\scriptsize Metrics follow Table~\ref{tab:overall_pi05}, with synchronous OpenVLA-OFT (OFT-Sync) as the reference for Residual Stall and Stall Reduction.\par}
  }
\end{table}

\subsection{End-to-End Control Quality and Boundary Stall} \label{sec:eval_e2e}

To answer Q1, Tables~\ref{tab:overall_pi05} and~\ref{tab:overall_oft} compare SR-VLA with synchronous execution and the latency-oriented method built on the same backbone in each family, reporting suite-wise and average task success alongside decision-boundary waiting time. Sharing the checkpoint and chunk length with SR-VLA, these baselines also cover the two classes in Section~\ref{sec:related_work}: VLASH increases $T_{\mathrm{hidden}}$ through future-state-aware asynchronous inference on $\pi_{0.5}$, and VLA-Cache shortens $T_{\mathrm{inf}}$ through temporal token reuse on OpenVLA-OFT. Stall Mean and Stall/Action report the mean wait per VLA query and per executed action, computed by averaging $T_{\mathrm{stall}}$ and $T_{\mathrm{stall}}/K$, respectively. Each table uses the synchronous baseline of its own backbone as the reference: Residual Stall normalizes Stall Mean by that of the reference, and Stall Reduction is the reference-to-method ratio of Stall/Action.

SR-VLA reduces decision-boundary waiting time while maintaining competitive task success rates across both backbone families. On $\pi_{0.5}$, it achieves a $2.67\times$ stall reduction over synchronous execution, with average success increasing from $96.4\%$ to $96.9\%$. The robot's idle share of each control cycle, $T_{\mathrm{stall}}/(T_{\mathrm{exec}}+T_{\mathrm{stall}})$, drops from $43.5\%$ to $22.3\%$. Against VLASH~\cite{vlash}, whose offset $d$ trades stall for success, SR-VLA matches $d{=}1$ in success with $1.95\times$ lower stall and beats $d{=}2$ on both. A larger $d{=}3$ cuts stall further only by dropping success below the synchronous baseline. On OpenVLA-OFT, SR-VLA achieves $1.27\times$ and $1.15\times$ stall reductions over the synchronous baseline and the synchronous approach with cache optimization~\cite{vlacache}, respectively.

Table~\ref{tab:suite_repair_summary} summarizes the suite-specific repair settings and statistics. Averaged equally across the four suites, the repair rate is $35.98\%$ and the all-query KV-update ratio is $20.48\%$, with unrepaired queries contributing zero updates. Dividing the latter by the repair rate gives the visual-token share rebuilt per triggered repair: $86.4\%$, $57.5\%$, $54.0\%$, and $34.6\%$ for Spatial, Object, Goal, and Long. This share closely follows $k_e$ over the 256 visual tokens per view, so the Top-$k$ error budget dominates the update set while the Top-ratio risk term adds a few task-relevant tokens.

\begin{table}[!t]
  \centering
\setlength{\abovecaptionskip}{3pt}
  \caption{Suite-wise repair settings and specifications.}
  \label{tab:suite_repair_summary}
  \footnotesize
  \setlength{\tabcolsep}{1.7pt}
\renewcommand{\arraystretch}{1.04}
  \begin{tabular*}{\columnwidth}{@{\extracolsep{\fill}}lccccccc@{}}
  \hline
  Suite & $\tau_{\mathrm{img}}$ & $\tau_p$ & $C_{\max}$ & \shortstack{Error $k_e$\\(per view)} & \shortstack{Risk $\alpha_r$\\(ratio)} & \shortstack{Repair\\rate (\%)} & \shortstack{Updated KV\\(all-q., \%)} \\
  \hline
  Spatial & 0.58 & 0.40 & 27 & 220 & 0.015 & 31.68 & 27.37 \\
  Object  & 1.07 & 0.59 & 17 & 147 & 0.049 & 43.27 & 24.88 \\
  Goal    & 0.57 & 0.47 & 29 & 137 & 0.018 & 30.09 & 16.24 \\
  Long    & 0.30 & 0.58 & 15 & 88  & 0.008 & 38.88 & 13.44 \\
  \hline
  \end{tabular*}
\end{table}

\subsection{Critical-Path Stall Breakdown} \label{sec:stall_breakdown}

To answer Q2, Figure~\ref{fig:latency_and_target_motion_results}(a) extends the LIBERO-Spatial ablations in Tables~\ref{tab:predictor_ablation} and~\ref{tab:repair_policy_ablation} and traces how each SR-VLA component moves the stall and success of $\pi_{0.5}$, averaged over the four suites. Directly starting inference asynchronously from launch-time inputs drops success to $62.4\%$. Adding the next-boundary predictor exposes only $2.74$~ms of stall and recovers success to $87.3\%$, so lookahead moves nearly all prediction off the critical path but leaves a success gap. The first repair level closes this gap: it validates every boundary against the real inputs and repairs only the triggered queries, reaching $96.8\%$ success at a mean stall of $60.46$~ms ($0.47\times$ of synchronous execution). The second level lowers the per-repair cost $\bar{T}_{\mathrm{rep}}$ in Eq.~(\ref{eq:real_path}), reducing stall to $47.95$~ms ($0.37\times$ of synchronous execution and $0.79\times$ of the first level alone) at $96.9\%$ success.

As the inset shows, exposed speculative computation ($T_{\mathrm{pred}}-T_{\mathrm{hidden}}$) accounts for only $2.74$~ms ($5.7\%$) of the $47.95$~ms mean stall, whereas VLASH ($d{=}2$) exposes its $60.25$~ms stall entirely as speculative computation. The remaining $45.21$~ms is real-input processing, which Eq.~(\ref{eq:real_path}) splits into validation, paid at every boundary ($28.1\%$ of the mean stall), and repair, paid only at the $35.98\%$ of queries that trigger it ($66.2\%$). After both levels, the residual stall lies on the real-input path rather than on prediction, making it the remaining target for runtime and accelerator optimization.

\subsection{Timing Robustness under Inference-Latency Inflation} \label{sec:delay_sensitivity}

VLA inference is slower on most deployment platforms than on this GPU and can further slow down with remote-inference overhead~\cite{rtc,vlash}. To answer Q3, we inject 10, 20, or 40~ms of additional latency by extending the VLA inference duration; for SR-VLA, the full delay is added to both speculative VLA inference and repair-stage VLA inference if repair is triggered.

Figure~\ref{fig:latency_and_target_motion_results}(b) shows that SR-VLA's stall grows more slowly with injected latency than those of synchronous $\pi_{0.5}$ and VLASH, both of which expose the added delay in full under the evaluated schedules. At the largest injection, SR-VLA exposes only $26.41$~ms of the added $40$~ms, for a mean stall of $74.36$~ms versus $100.25$~ms for VLASH ($d{=}2$). Because $T_{\mathrm{pred}}$ overlaps ongoing action execution, the execution window absorbs part of the added latency.

\subsection{Boundary Validation under Environment Dynamics} \label{sec:env_dynamics}

Moving a task-relevant object during the current action chunk invalidates speculative state prepared before the real next-boundary inputs. To answer Q4, we test the complete lookahead-and-repair path on LIBERO-Spatial with the frozen $\pi_{0.5}$ backbone and $K=5$ by displacing the target object by 0, 1, 3, or 5~cm, where 0~cm denotes no motion. The displacement is applied before action 1, 3, or 5, chosen at random to cover the beginning, middle, and end of the chunk.

Figure~\ref{fig:latency_and_target_motion_results}(c) shows that success decreases with displacement for all three methods, while SR-VLA tracks synchronous $\pi_{0.5}$: both decline from $97.8\%$ at 0~cm to $72.2\%$ at 5~cm. Its advantage over VLASH peaks at 3~cm, where it matches the synchronous reference.

Larger displacements trigger more frequent repair: the repair rate rises from $31.68\%$ at 0~cm to $60.26\%$ at 5~cm. SR-VLA still achieves a $1.93\times$ stall reduction over synchronous $\pi_{0.5}$ at 1~cm and retains $1.52\times$ at 5~cm, so boundary validation responds to real mismatches while preserving most of the stall reduction.

\section{Conclusion}

SR-VLA combines predictive lookahead during action execution with real-input-guided validation and selective KV-cache repair to reduce decision-boundary waiting. On LIBERO with frozen $\pi_{0.5}$, it reduces mean exposed stall from $128.11$ to $47.95$~ms, a $2.67\times$ stall reduction over synchronous execution, and on OpenVLA-OFT it achieves $1.27\times$ and $1.15\times$ stall reductions over synchronous OpenVLA-OFT and VLA-Cache, respectively. It also slows stall growth under injected inference latency and keeps synchronous-level success under target motion with a $1.52\times$--$1.93\times$ stall reduction. The remaining stall lies on the real-input path of validation and repair, the next target for runtime and accelerator support.

\clearpage
\renewcommand*{\bibfont}{\footnotesize}
\printbibliography

@inproceedings{act,
  author    = {Zhao, Tony Z. and Kumar, Vikash and Levine, Sergey and Finn, Chelsea},
  title     = {Learning Fine-Grained Bimanual Manipulation with Low-Cost Hardware},
  booktitle = {Proceedings of Robotics: Science and Systems (RSS)},
  year      = {2023}, 
  month     = {7},
  address   = {Daegu, Republic of Korea}, 
}

@inproceedings{draftverify,
  author    = {Zhang, Jun and Wang, Jun and Li, Huan and Shou, Lidan and Chen, Ke and Chen, Gang and Mehrotra, Sharad},
  title     = {Draft \& Verify: Lossless Large Language Model Acceleration via Self-Speculative Decoding},
  booktitle = {Proceedings of the Annual Meeting of the Association for Computational Linguistics (ACL)},
  year      = {2024},
  month     = {8},
  address   = {Bangkok, Thailand},
}

@inproceedings{dyslvla,
  author        = {Yang, Zebin and Qi, Yijiahao and Xie, Tong and Yu, Bo and Liu, Shaoshan and Li, Meng},
  title         = {{DySL-VLA}: Efficient Vision-Language-Action Model Inference via Dynamic-Static Layer-Skipping for Robot Manipulation},
  booktitle = {Design Automation Conference (DAC)},
  year      = {2026},
  month     = {7},
  address   = {Long Beach, CA, USA},
}

@inproceedings{efficientvla,
  author    = {Yang, Yantai and Wang, Yuhao and Wen, Zichen and Luo, Zhongwei and Zou, Chang and Zhang, Zhipeng and Wen, Chuan and Zhang, Linfeng},
  title     = {{EfficientVLA}: Training-Free Acceleration and Compression for Vision-Language-Action Models},
  booktitle = {The 39th Annual Conference on Neural Information Processing Systems (NeurIPS)},
  year      = {2025},
  month     = {12},
  address   = {San Diego, CA, USA},
}

@article{smolvla,
  title   = {{SmolVLA}: A Vision-Language-Action Model for Affordable and Efficient Robotics},
  author  = {Shukor, Mustafa and Aubakirova, Dana and Capuano, Francesco and Kooijmans, Pepijn and Palma, Steven and Zouitine, Adil and Aractingi, Michel and Pascal, Caroline and Russi, Martino and Marafioti, Andres and Alibert, Simon and Cord, Matthieu and Wolf, Thomas and Cadene, Remi},
  journal = {arXiv preprint arXiv:2506.01844},
  year    = {2025}
}

@article{flashvla,
  title   = {Think Twice, Act Once: Token-Aware Compression and Action Reuse for Efficient Inference in Vision-Language-Action Models},
  author  = {Tan, Xudong and Yang, Yaoxin and Ye, Peng and Zheng, Jialin
             and Bai, Bizhe and Wang, Xinyi and Hao, Jia and Chen, Tao},
  journal = {arXiv preprint arXiv:2505.21200},
  year    = {2025}
}

@inproceedings{specvla,
  title     = {{Spec-VLA}: Speculative Decoding for Vision-Language-Action Models with Relaxed Acceptance},
  author    = {Wang, Songsheng and Yu, Rucheng and Yuan, Zhihang and Yu, Chao and Gao, Feng and Wang, Yu and Wong, Derek F.},
  booktitle = {Proceedings of the 2025 Conference on Empirical Methods in Natural Language Processing (EMNLP)},
  year      = {2025},
  month     = {11},
  address   = {Suzhou, China},
}

@article{vlash,
  title   = {{VLASH}: Real-Time {VLA}s via Future-State-Aware Asynchronous Inference},
  author  = {Tang, Jiaming and Sun, Yufei and Zhao, Yilong and Yang, Shang and Lin, Yujun and Zhang, Zhuoyang and Hou, James and Lu, Yao and Liu, Zhijian and Han, Song},
  journal = {arXiv preprint arXiv:2512.01031},
  year    = {2025}
}

@inproceedings{gr1,
  author    = {Wu, Hongtao and Jing, Ya and Cheang, Chilam and Chen, Guangzeng and Xu, Jiafeng and Li, Xinghang and Liu, Minghuan and Li, Hang and Kong, Tao},
  title     = {Unleashing Large-Scale Video Generative Pre-training for Visual Robot Manipulation},
  booktitle = {The 12th International Conference on Learning Representations (ICLR)},
  year      = {2024},
  month     = {5},
  address   = {Vienna, Austria},
}

@inproceedings{kerv,
  author    = {Zheng, Zhihao and Mao, Zhihao and Li, Maoliang and Chen, Jiayu and Sun, Xinhao and Zhang, Zhaobo and Cao, Donggang and Mei, Hong and Chen, Xiang},
  title     = {{KERV}: Kinematic-Rectified Speculative Decoding for Embodied VLA Models},
  booktitle = {Design Automation Conference (DAC)},
  year      = {2026},
  month     = {7},
  address   = {Long Beach, CA, USA},
}

@inproceedings{actionflow,
  author    = {Dai, Yuntao and Gu, Hang and Wang, Teng and Cheng, Qianyu and Zheng, Yifei and Qiu, Zhiyong and Gong, Lei and Lou, Wenqi and Zhou, Xuehai},
  title     = {{ActionFlow}: A Pipelined Action Acceleration for Vision Language Models on Edge},
  booktitle = {Design Automation Conference (DAC)},
  year      = {2026},
  month     = {7},
  address   = {Long Beach, CA, USA},
}

@inproceedings{keep,
  author    = {Yang, Zebin and Xie, Tong and Lu, Baotong and Liu, Shaoshan and Yu, Bo and Li, Meng},
  title     = {{KEEP}: A KV-Cache-Centric Memory Management System for Efficient Embodied Planning},
  booktitle = {Design Automation Conference (DAC)},
  year      = {2026},
  month     = {7},
  address   = {Long Beach, CA, USA},
}

@inproceedings{libero,
  author    = {Liu, Bo and Zhu, Yifeng and Gao, Chongkai and Feng, Yihao and Liu, Qiang and Zhu, Yuke and Stone, Peter},
  title     = {{LIBERO}: Benchmarking Knowledge Transfer for Lifelong Robot Learning}, 
  booktitle = {The 37th Conference on Neural Information Processing Systems (NeurIPS)},
  year      = {2023},
  month     = {12},
  address   = {New Orleans, LA, USA},
}

@inproceedings{openvla,
  author    = {Kim, Moo Jin and Pertsch, Karl and Karamcheti, Siddharth and Xiao, Ted and Balakrishna, Ashwin and Nair, Suraj and Rafailov, Rafael and Foster, Ethan P and Sanketi, Pannag R and Vuong, Quan and Kollar, Thomas and Burchfiel, Benjamin and Tedrake, Russ and Sadigh, Dorsa and Levine, Sergey and Liang, Percy and Finn, Chelsea},
  title     = {{OpenVLA}: An Open-Source Vision-Language-Action Model},
  booktitle = {Proceedings of The 9th Annual Conference on Robot Learning (CoRL)},
  year      = {2025},
  month     = {9},
  address   = {Seoul, South Korea},
}

@inproceedings{openvlaoft,
  author    = {Kim, Moo Jin and Finn, Chelsea and Liang, Percy},
  title     = {Fine-Tuning Vision-Language-Action Models: Optimizing Speed and Success},
  booktitle = {Proceedings of Robotics: Science and Systems (RSS)},
  year      = {2025},
  month     = {6},
  address   = {Los Angeles, CA, USA},
}

@inproceedings{parallelvla,
  author    = {Song, Wenxuan and Chen, Jiayi and Ding, Pengxiang and Zhao, Han and Zhao, Wei and Zhong, Zhide and Ge, Zongyuan and Li, Zhijun and Wang, Donglin and Wang, Lujia and Ma, Jun and Li, Haoang},
  title     = {{PD-VLA}: Accelerating Vision-Language-Action Model Integrated with Action Chunking via Parallel Decoding},
  booktitle = {2025 IEEE/RSJ International Conference on Intelligent Robots and Systems (IROS)},
  year      = {2025},
  month     = {10},
  address   = {Hangzhou, China},
}

@inproceedings{pi0,
  author    = {Black, Kevin and Brown, Noah and Driess, Danny and Esmail, Adnan and Equi, Michael and Finn, Chelsea and Fusai, Niccolo and Groom, Lachy and Hausman, Karol and Ichter, Brian and Jakubczak, Szymon and Jones, Tim and Ke, Liyiming and Levine, Sergey and Li-Bell, Adrian and Mothukuri, Mohith and Nair, Suraj and Pertsch, Karl and Shi, Lucy and Smith, Laura and Tanner, James and Vuong, Quan and Walling, Anna and Wang, Haohuan and Zhilinsky, Ury},
  title     = {{$\pi_0$}: A Vision-Language-Action Flow Model for General Robot Control},
  booktitle = {Proceedings of Robotics: Science and Systems (RSS)},
  year      = {2025},
  month     = {6},
  address   = {Los Angeles, CA, USA},
}

@inproceedings{pi05,
  author    = {Black, Kevin and Brown, Noah and Darpinian, James and Dhabalia, Karan and Driess, Danny and Esmail, Adnan and Equi, Michael and Finn, Chelsea and Fusai, Niccolo and Galliker, Manuel Y. and Ghosh, Dibya and Groom, Lachy and Hausman, Karol and Ichter, Brian and Jakubczak, Szymon and Jones, Tim and Ke, Liyiming and LeBlanc, Devin and Levine, Sergey and Li-Bell, Adrian and Mothukuri, Mohith and Nair, Suraj and Pertsch, Karl and Ren, Allen Z. and Shi, Lucy Xiaoyang and Smith, Laura and Springenberg, Jost Tobias and Stachowicz, Kyle and Tanner, James and Vuong, Quan and Walke, Homer and Walling, Anna and Wang, Haohuan and Yu, Lili and Zhilinsky, Ury},
  title     = {{$\pi_{0.5}$}: A Vision-Language-Action Model with Open-World Generalization},
  booktitle = {Proceedings of The 9th Annual Conference on Robot Learning (CoRL)},
  year      = {2025},
  month     = {9},
  address   = {Seoul, South Korea},
}

@misc{qail,
  author        = {Park, Seongmin and Kim, Hyungmin and Jeon, Wonseok and Yang, Juyoung and Jeon, Byeongwook and Oh, Yoonseon and Choi, Jungwook},
  title         = {Quantization-Aware Imitation-Learning for Resource-Efficient Robotic Control},
  year          = {2024},
  note   = {arXiv:2412.01034}
}

@inproceedings{robodreamer,
  author    = {Zhou, Siyuan and Du, Yilun and Chen, Jiaben and Li, Yandong and Yeung, Dit-Yan and Gan, Chuang},
  title     = {{RoboDreamer}: Learning Compositional World Models for Robot Imagination},
  booktitle = {Proceedings of the 41st International Conference on Machine Learning (ICML)},
  year      = {2024},
  month     = {7},
  address   = {Vienna, Austria},
}

@inproceedings{rt2,
  author    = {Zitkovich, Brianna and Yu, Tianhe and Xu, Sichun and Xu, Peng and Xiao, Ted and Xia, Fei and Wu, Jialin and Wohlhart, Paul and Welker, Stefan and Wahid, Ayzaan and Vuong, Quan and Vanhoucke, Vincent and Tran, Huong and Soricut, Radu and Singh, Anikait and Singh, Jaspiar and Sermanet, Pierre and Sanketi, Pannag R. and Salazar, Grecia and Ryoo, Michael S. and Reymann, Krista and Rao, Kanishka and Pertsch, Karl and Mordatch, Igor and Michalewski, Henryk and Lu, Yao and Levine, Sergey and Lee, Lisa and Lee, Tsang-Wei Edward and Leal, Isabel and Kuang, Yuheng and Kalashnikov, Dmitry and Julian, Ryan and Joshi, Nikhil J. and Irpan, Alex and Ichter, Brian and Hsu, Jasmine and Herzog, Alexander and Hausman, Karol and Gopalakrishnan, Keerthana and Fu, Chuyuan and Florence, Pete and Finn, Chelsea and Dubey, Kumar Avinava and Driess, Danny and Ding, Tianli and Choromanski, Krzysztof Marcin and Chen, Xi and Chebotar, Yevgen and Carbajal, Justice and Brown, Noah and Brohan, Anthony and Arenas, Montserrat Gonzalez and Han, Kehang},
  title     = {{RT-2}: Vision-Language-Action Models Transfer Web Knowledge to Robotic Control},
  booktitle = {Proceedings of The 7th Conference on Robot Learning (CoRL)},
  year      = {2023},
  month     = {11},
  address   = {Atlanta, GA, USA},
}

@inproceedings{rtc,
  author    = {Black, Kevin and Galliker, Noah Y. and Levine, Sergey},
  title     = {Real-Time Execution of Action Chunking Flow Policies},
  booktitle = {The 39th Annual Conference on Neural Information Processing Systems (NeurIPS)},
  year      = {2025},
  month     = {12},
  address   = {San Diego, CA, USA},
}

@inproceedings{valuepred,
  author    = {Lipasti, Mikko H. and Wilkerson, Christopher B. and Shen, John Paul},
  title     = {Value Locality and Load Value Prediction},
  booktitle = {Proceedings of the 7th International Conference on Architectural Support for Programming Languages and Operating Systems (ASPLOS)},
  year      = {1996},
  month     = {10},
  address   = {Cambridge, MA, USA},
}

@inproceedings{specdec,
  author    = {Leviathan, Yaniv and Kalman, Matan and Matias, Yossi},
  title     = {Fast Inference from Transformers via Speculative Decoding},
  booktitle = {Proceedings of the 40th International Conference on Machine Learning (ICML)},
  year      = {2023},
  month     = {7},
  address   = {Honolulu, HI, USA},
}

@inproceedings{spvla,
  author    = {Li, Ye and Meng, Yuan and Sun, Zewen and Ji, Kangye and Tang, Chen and Fan, Jiajun and Ma, Xinzhu and Xia, Shutao and Wang, Zhi and Zhu, Wenwu},
  title     = {{SP-VLA}: A Joint Model Scheduling and Token Pruning Approach for VLA Model Acceleration},
  booktitle = {The 14th International Conference on Learning Representations (ICLR)},
  year      = {2026},
  month     = {4},
  address   = {Rio de Janeiro, Brazil},
}

@article{tinyvla,
  author  = {Wen, Junjie and Zhu, Yichen and Li, Jinming and Zhu, Minjie and Tang, Zhibin and Wu, Kun and Xu, Zhiyuan and Liu, Ning and Cheng, Ran and Shen, Chaomin and Peng, Yaxin and Feng, Feifei and Tang, Jian},
  title   = {{TinyVLA}: Toward Fast, Data-Efficient Vision-Language-Action Models for Robotic Manipulation},
  journal = {IEEE Robotics and Automation Letters},
  volume  = {10},
  number  = {4},
  pages   = {3988--3995},
  year    = {2025},
  month   = {4},
}

@inproceedings{vlacache,
  author    = {Xu, Siyu and Wang, Yunke and Xia, Chenghao and Zhu, Dihao and Huang, Tao and Xu, Chang},
  title     = {{VLA-Cache}: Efficient Vision-Language-Action Manipulation via Adaptive Token Caching},
  booktitle = {The 39th Annual Conference on Neural Information Processing Systems (NeurIPS)},
  year      = {2025},
  month     = {12},
  address   = {San Diego, CA, USA},
}

@inproceedings{cogvla,
  author    = {Li, Wei and Zhang, Renshan and Shao, Rui and He, Jie and Nie, Liqiang},
  title     = {{CogVLA}: Cognition-Aligned Vision-Language-Action Models via Instruction-Driven Routing \& Sparsification},
  booktitle = {The 39th Annual Conference on Neural Information Processing Systems (NeurIPS)},
  year      = {2025},
  month     = {12},
  address   = {San Diego, CA, USA},
}

@inproceedings{specprunevla,
  author    = {Wang, Hanzhen and Xu, Jiaming and Xiang, Yushun and Pan, Jiayi and Zhou, Yongkang and Li, Yong-Lu and Dai, Guohao},
  title     = {{SpecPrune-VLA}: Accelerating Vision-Language-Action Models via Action-Aware Self-Speculative Pruning},  
  booktitle = {Proceedings of the 43rd International Conference on Machine Learning (ICML)},
  year      = {2026},
  month     = {7},
  address   = {Seoul, South Korea},
}

@article{jetsonpi,
  author        = {Yang, Zebin and Wang, Qi and Wang, Yunhe and Guo, Xiurui and Yu, Bo and Liu, Shaoshan and Xu, Jiafeng and Dong, Hao and Li, Meng},
  title         = {{EagleVLA}: Towards Onboard Real-Time Robot Control via Foresight-Aligned Asynchronous Inference},
  journal       = {arXiv preprint arXiv:2607.12659},
  year          = {2026},
}

@article{futurertc,
  author        = {Jiang, Hai and Zou, Yixian and Liang, Binbin and Liu, Boqian and Meng, Fanman and Liu, Shuaicheng},
  title         = {{FutureRTC}: Real-Time Robot Execution with Anticipatory-Conditioned Action Chunking},
  journal       = {arXiv preprint arXiv:2607.24008},
  year          = {2026},
}

@article{pactvla,
  author        = {Zhu, Yixiang and Wu, Ruilong and Chen, Yonghao and Zou, Jiaxiang and Hu, Yusong and Chen, Xinyu},
  title         = {{PACT-VLA}: Speculative Action-Chunk Verification with Pre-Commit Observations},
  journal       = {arXiv preprint arXiv:2610.28019},
  year          = {2026},
}

@article{checkvla,
  author        = {Liu, Yushan and Sun, Peibo and Chao, Xintao and others},
  title         = {{CheckVLA}: Execution-Time Verification with Action-Conditioned World Model for Long-Horizon Mobile Manipulation},
  journal       = {arXiv preprint arXiv:2607.26789},
  year          = {2026},
}

@article{realtimeflash,
  author        = {Niu, Jiahui and Gu, Kefan and Zhao, Yucheng and Liang, Shengwen and Wang, Tiancai and Hu, Xing and Wang, Ying and Li, Huawei},
  title         = {{Realtime-VLA FLASH}: Speculative Inference Framework for Diffusion-based {VLA}s},
  journal       = {arXiv preprint arXiv:2605.13778},
  year          = {2026},
}

@inproceedings{cacheblend,
  author    = {Yao, Jiayi and Li, Hanchen and Liu, Yuhan and Ray, Siddhant and Cheng, Yihua and Zhang, Qizheng and Du, Kuntai and Lu, Shan and Jiang, Junchen},
  title     = {{CacheBlend}: Fast Large Language Model Serving for {RAG} with Cached Knowledge Fusion},
  booktitle = {Proceedings of the 20th European Conference on Computer Systems (EuroSys)},
  year      = {2025},
  month     = {3},
  address   = {Rotterdam, Netherlands},
}
\end{document}